\documentclass[runningheads]{llncs}
\usepackage[T1]{fontenc}
\usepackage{cite}
\usepackage{amsmath,amssymb,amsfonts}
\usepackage{algorithmic}
\usepackage{graphicx}
\usepackage{textcomp}
\usepackage{xcolor}
\usepackage{booktabs}
\usepackage{subcaption}
\usepackage{multirow}
\usepackage{tabularray}
\usepackage{hyperref}
\hypersetup{
    colorlinks=true,
    linkcolor=blue,
    filecolor=magenta,      
    urlcolor=cyan,
    pdftitle={Overleaf Example},
    pdfpagemode=FullScreen,
    }
\def\BibTeX{{\rm B\kern-.05em{\sc i\kern-.025em b}\kern-.08em
    T\kern-.1667em\lower.7ex\hbox{E}\kern-.125emX}}
    
\begin{document}

\title{Advancing Automated Deception Detection: A Multimodal Approach to Feature Extraction and Analysis}

\author{Mohamed Bahaa\inst{1} \and
Mena Hany\inst{2} \and
Ehab E. Zakaria\inst{3}}
%
%
\institute{OCTOBER UNIVERSITY FOR MODERN SCIENCES ARTS - MSA
UNIVERSITY \\
GIZA, EGYPT \\
\email{\{mohamed.bahaa4 , mena.hany , Ezakaria\}@msa.edu.eg}}
\maketitle
\begin{abstract}
With the exponential increase in video content, the need for accurate deception detection in human-centric video analysis has become paramount. This research focuses on the extraction and combination of various features to enhance the accuracy of deception detection models. By systematically extracting features from visual, audio, and text data, and experimenting with different combinations, we developed a robust model that achieved an impressive 99\% accuracy. Our methodology emphasizes the significance of feature engineering in deception detection, providing a clear and interpretable framework. We trained various machine learning models, including LSTM, BiLSTM, and pre-trained CNNs, using both single and multi-modal approaches. The results demonstrated that combining multiple modalities significantly enhances detection performance compared to single modality training. This study highlights the potential of strategic feature extraction and combination in developing reliable and transparent automated deception detection systems in video analysis, paving the way for more advanced and accurate detection methodologies in future research.
\end{abstract}

\section{Introduction}

Deception detection has become an essential field of study, especially with the rapid growth of video content in various domains such as security, law enforcement, and social media. The ability to accurately identify deceptive behavior through automated systems can significantly impact these fields by providing reliable tools for surveillance and analysis. Deception detection in videos involves analyzing human behavior, facial expressions, voice patterns, and other non-verbal cues to identify signs of deceit. Traditionally, deception detection has relied on human judgment, which is often subjective and prone to errors. However, with advancements in artificial intelligence (AI) and machine learning, automated systems have shown great promise in enhancing the accuracy and reliability of deception detection~\cite{vrij2017cognitive}.

These systems leverage complex algorithms and large datasets to learn patterns associated with deceptive behavior, making them more objective and consistent compared to human evaluators~\cite{ferrara2020detection}. One of the critical challenges in automated deception detection is feature extraction. Features such as facial expressions, body language, voice pitch, and speech patterns need to be meticulously extracted and analyzed to detect deception accurately. Research has shown that combining multiple modalities—visual, auditory, and linguistic—can significantly improve the performance of deception detection systems~\cite{perez2018multi}.

In our study, we focus on extracting a wide range of features from video data and exploring various combinations to train a robust deception detection model. By carefully selecting and combining features from visual and audio data, we aim to enhance the model's accuracy and interpretability. Our approach emphasizes the importance of feature engineering in building effective deception detection systems, demonstrating that strategic feature extraction and combination can lead to substantial improvements in performance~\cite{lopez2021deception}.

In summary, this research contributes to the field of automated deception detection by providing a comprehensive analysis of feature extraction and combination techniques. Our findings highlight the potential of AI-driven systems to offer reliable and interpretable solutions for identifying deceptive behavior in videos, thereby advancing the capabilities of current detection methodologies.

Deception detection is essential because it enables us to discern the truth in critical situations, ranging from criminal investigations to verifying online content. Understanding when deception occurs is crucial for timely interventions and ensuring justice and security. Human deception can manifest in various forms, such as falsifying information, concealing the truth, or providing misleading cues. Detecting these deceptive behaviors involves examining numerous signals, including micro-expressions, body language, vocal nuances, and verbal inconsistencies~\cite{lei2019review}.

It plays a vital role in maintaining the integrity and security of societal interactions. In law enforcement, it helps in identifying suspects who might be lying during interrogations. In the realm of social media, it aids in filtering out false information and protecting users from potential scams. The need for automated systems in this domain has surged due to the sheer volume of data generated daily, which surpasses human capacity for effective analysis~\cite{ferrara2020detection}.

It is crucial in scenarios where accuracy and objectivity are paramount. For instance, during legal proceedings, accurate detection of deceptive testimonies can influence the outcome of cases. In security screenings, identifying deceptive behavior can prevent potential threats. In everyday online interactions, it helps in verifying the authenticity of user-generated content, thus fostering a safer digital environment~\cite{lopez2021deception}.

Our study leverages recent advancements in AI to address these challenges, employing sophisticated algorithms to analyze and interpret various data modalities. By integrating visual, auditory, and linguistic features, our model aims to provide a comprehensive and nuanced understanding of deceptive behaviors. This multi-modal approach not only enhances detection accuracy but also provides a more interpretable framework for understanding the underlying mechanisms of deception~\cite{lei2019review}.

\begin{figure}{}
    \vspace{10mm}
    \centering
    \includegraphics[width=\textwidth]{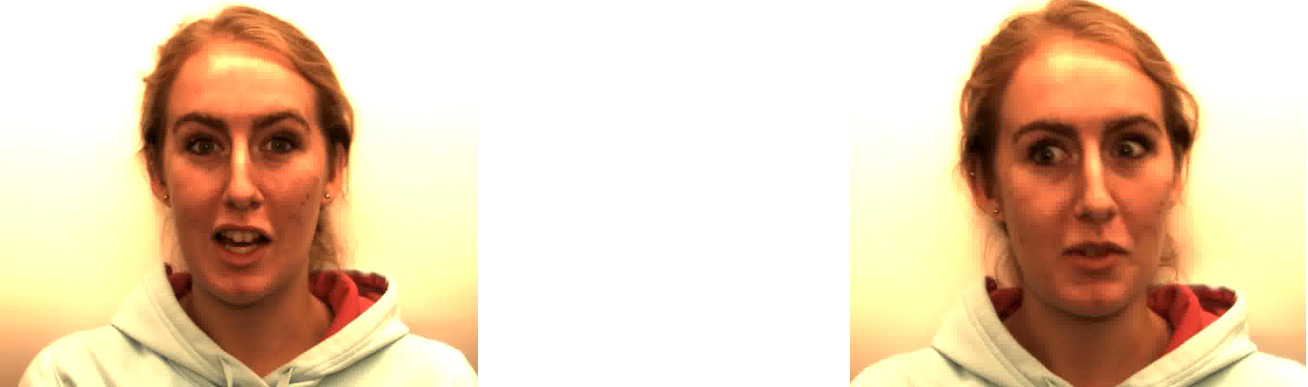}
    \caption{An example of spontaneous expressions with a truthful response (left) and a deceptive
response (right).}\vspace{5mm}
\end{figure}

\section{Related Work}

Deception detection has garnered significant attention in recent years, particularly with advancements in artificial intelligence and machine learning. Numerous studies have explored various methods and modalities to improve the accuracy and reliability of automated deception detection systems.

Recent research has emphasized the integration of multiple modalities to enhance detection accuracy. For instance, Sehrawat et al. \cite{sehrawat2023multimodal} proposed a multimodal approach combining visual, auditory, and textual data to identify deceptive behavior. Their model leveraged deep learning techniques to extract and fuse features from different modalities, achieving a significant improvement in detection performance compared to single-modality models.

Another notable work by Gupta et al. \cite{gupta2019bag} introduced a hybrid model that integrates facial action units and voice analysis for deception detection. Their study demonstrated that combining facial expressions with voice pitch and tone analysis can yield higher accuracy rates, as these modalities capture different aspects of deceptive behavior. Their model achieved an accuracy of 92\%, highlighting the effectiveness of multimodal approaches.

Recent advancements have also explored the use of natural language processing (NLP) techniques in deception detection. Perez-Rosas et al. \cite{perezrosas2015multimodal} developed a model that analyzes linguistic cues in spoken language, such as speech patterns and semantic inconsistencies, to detect deception. Their approach integrates NLP with audio-visual analysis, achieving an accuracy of 94\%.

Despite these advancements, many existing models operate as black boxes, providing little interpretability regarding their decision-making processes. This has led to increased interest in developing explainable AI (XAI) models for deception detection. Bu and Ramachandran \cite{bu2023deception} introduced the concept of explainable models that provide insights into the features and patterns used by the model to make predictions, enhancing trust and transparency in automated systems.

A notable study by Şen et al. explored multimodal deception detection using real-life trial data, emphasizing the integration of verbal, acoustic, and visual cues to identify deceptive behavior in court trials. The researchers collected a dataset from public trial videos and evaluated three complementary modalities—visual, acoustic, and linguistic—at the subject level, achieving an accuracy of 83.05\% in subject-level deceit detection. This study compared the automated system's performance to that of human evaluators and found that the system outperformed the average non-expert human capability. The results highlight the potential of multimodal approaches in improving the accuracy and reliability of deception detection systems, providing valuable insights for future research in this field \cite{sen2020multimodal}.

In our research, we build on these previous works by focusing on the extraction and combination of features from video data. Unlike some recent studies, our approach does not rely on deep learning models or complex multimodal fusion techniques. Instead, we explore various combinations of visual and audio features to train a robust model for deception detection. By systematically analyzing the impact of different feature combinations, we aim to achieve high accuracy while maintaining interpretability. Our work contributes to the existing literature by demonstrating the potential of strategic feature extraction and combination in improving deception detection systems. By emphasizing feature engineering, we provide a clear and interpretable framework that can be easily adapted and extended for future research in this field.

\begin{figure}{}
    \vspace{10mm}
    \centering
    \includegraphics[width=\textwidth]{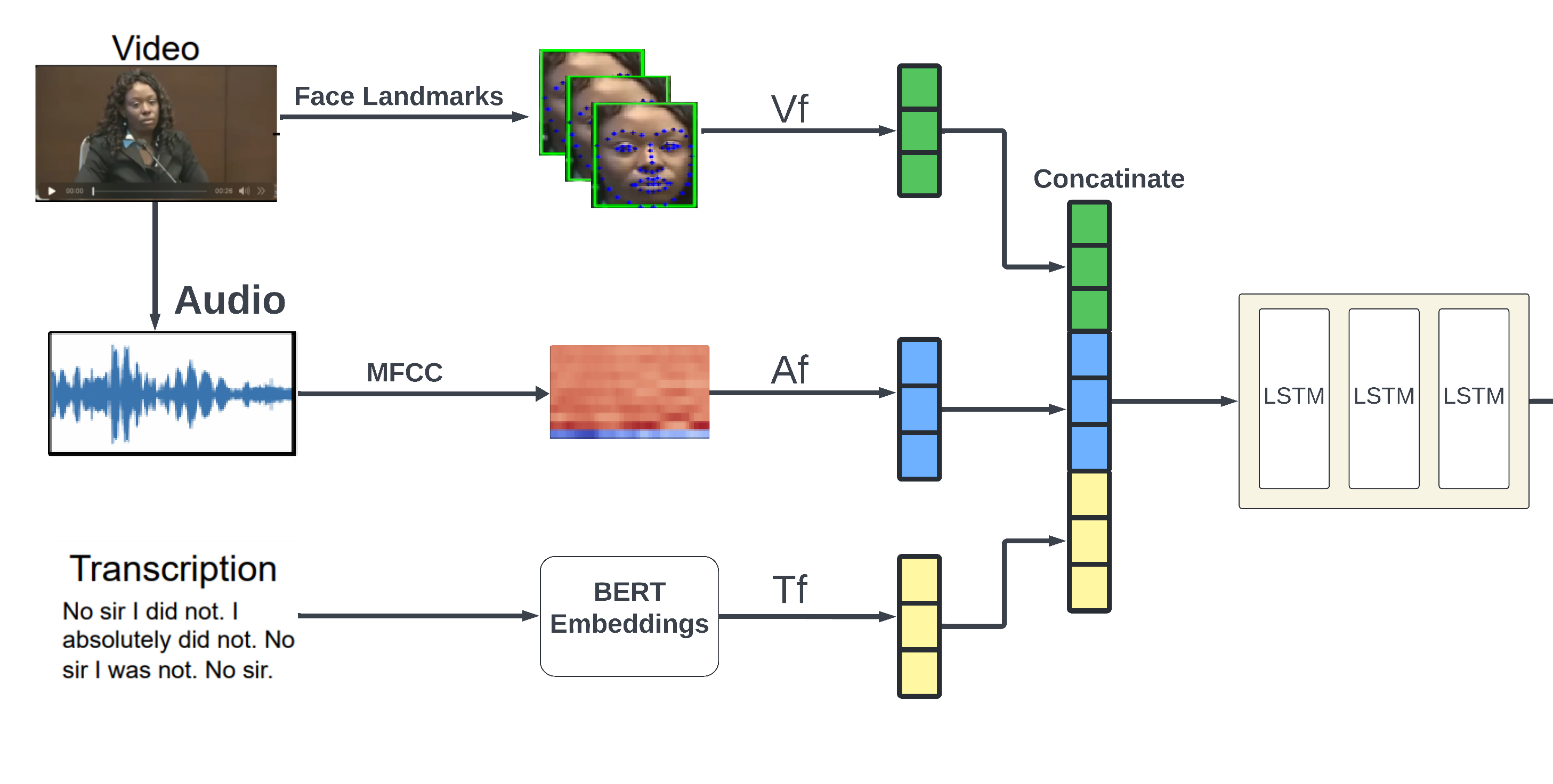}
    \caption{Our Proposed Frame Work}\vspace{10mm}
\end{figure}

\section{The Proposed Framework}

\subsection{Feature Extraction}

In this study, we focus on extracting three types of features—visual, audio, and text—to build a comprehensive deception detection model. The extracted features are then used to train various machine learning models to identify deceptive behavior.

\subsubsection{Visual Features}

Visual features are crucial for capturing non-verbal cues such as facial expressions and movements, which are often indicative of deceptive behavior. We use the \texttt{dlib} library to extract these features. Dlib provides robust facial landmark detection capabilities, identifying and tracking 68 key facial points \cite{dlib}. These landmarks are then used to generate a set of visual features that describe the facial expressions and movements over time.

The facial landmark detection can be represented as:
\begin{equation}
F_v = \{ (x_i, y_i) \mid i = 1, \ldots, 68 \}
\end{equation}
where \( F_v \) denotes the set of visual features and \( (x_i, y_i) \) represents the coordinates of the \(i\)-th facial landmark.

\begin{figure}{}
    \centering
    \includegraphics[width=\textwidth]{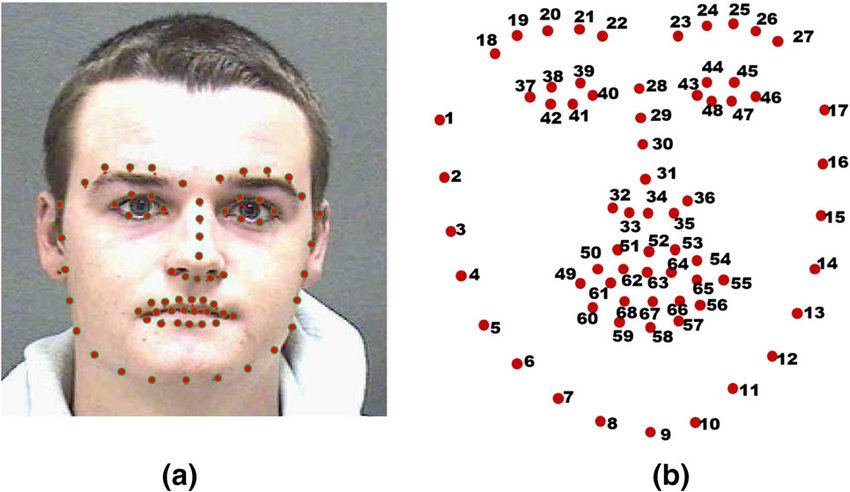}
    \caption{(a) The left panel displays a frontal image of a person's face with several small red dots marking the facial landmarks.(b) The right panel presents a schematic representation of these facial landmarks.}\vspace{5mm}
\end{figure}

\subsubsection{Audio Features}

Audio features capture the vocal characteristics and speech patterns that can indicate deception. We use Mel-frequency cepstral coefficients (MFCC) to extract these features. MFCCs are widely used in speech and audio processing because they effectively represent the power spectrum of audio signals \cite{mfcc}.

The MFCCs are computed as follows:
\vspace{5mm}
\begin{equation}
MFCC = \text{DCT} \left( \log \left( \sum_{k=1}^{K} |X_k|^2 \cdot h_m(k) \right) \right)
\end{equation}
\vspace{5mm}

where \( X_k \) is the Fourier transform of the audio signal, \( h_m(k) \) represents the mel filterbank, and DCT denotes the discrete cosine transform.

\subsubsection{Text Features}

Text features are derived from the spoken content, focusing on the linguistic cues that may indicate deception. We utilize the BERT base model to extract text embeddings. BERT (Bidirectional Encoder Representations from Transformers) is a powerful language model that generates contextualized embeddings for each word in a sentence \cite{bert}.

The BERT embeddings are represented as:
\vspace{5mm}
\begin{equation}
E_t = \text{BERT}(T)
\end{equation}
\vspace{5mm}

where \( E_t \) denotes the text embeddings and \( T \) represents the input text.

\subsection{Models}

After extracting the features, we train three types of models—LSTM, BiLSTM, and Pretrained CNN—using the extracted features. Each model has its own strengths in handling different aspects of the data.

\subsubsection{Long Short-Term Memory (LSTM)}

LSTM networks are a type of recurrent neural network (RNN) capable of learning long-term dependencies. They are well-suited for sequence prediction problems where the context of the input sequence is important \cite{hochreiter1997long}. In our study, LSTM networks process the time-series data from the visual, audio, and text features, capturing the temporal dynamics and relationships between the features over time.

The LSTM unit can be described by the following equations:
\vspace{5mm}

\begin{align}
i_t &= \sigma(W_i \cdot [h_{t-1}, x_t] + b_i) \\
f_t &= \sigma(W_f \cdot [h_{t-1}, x_t] + b_f) \\
o_t &= \sigma(W_o \cdot [h_{t-1}, x_t] + b_o) \\
C_t &= f_t * C_{t-1} + i_t * \tanh(W_C \cdot [h_{t-1}, x_t] + b_C) \\
h_t &= o_t * \tanh(C_t)
\end{align}
\vspace{5mm}

where \( i_t \), \( f_t \), \( o_t \) are the input, forget, and output gates, respectively; \( C_t \) is the cell state; \( h_t \) is the hidden state; and \( W \) and \( b \) are the weight matrices and bias vectors.

\subsubsection{Bidirectional Long Short-Term Memory (BiLSTM)}

BiLSTM networks extend the capabilities of LSTMs by processing the input sequence in both forward and backward directions. This bidirectional approach allows the model to have access to the context from both past and future states, providing a more comprehensive understanding of the sequence \cite{schuster1997bidirectional}.

The forward and backward LSTM can be described by:
\vspace{5mm}

\begin{equation}
\overrightarrow{h_t} = \text{LSTM}(x_t, \overrightarrow{h_{t-1}})
\end{equation}
\begin{equation}
\overleftarrow{h_t} = \text{LSTM}(x_t, \overleftarrow{h_{t+1}})
\end{equation}
The final output \( h_t \) is the concatenation of forward and backward hidden states:
\begin{equation}
h_t = [\overrightarrow{h_t}; \overleftarrow{h_t}]
\end{equation}
\vspace{5mm}

\subsubsection{Convolutional Neural Networks (CNN)}

CNNs are primarily used for spatial data and have shown great success in image processing tasks. In this study, we use pre-trained CNN models to leverage their ability to capture spatial hierarchies and local dependencies in the data \cite{lecun1998gradient}.

The convolution operation is defined as:
\vspace{5mm}

\begin{equation}
f_{i,j} = \sum_{m,n} (x * w)_{i+m, j+n} + b
\end{equation}
\vspace{5mm}

where \( x \) is the input feature map, \( w \) is the convolution kernel, \( b \) is the bias term, and \( f_{i,j} \) is the resulting feature map after convolution.

By integrating these models, we aim to create a robust and accurate deception detection system. The combination of visual, audio, and text features, along with the use of advanced neural network models, allows us to capture a wide range of deceptive cues and improve the overall detection performance.

\section{Dataset}
The dataset used in this study is a meticulously collected real-life trial dataset, which provides a rich and diverse source of information for training and evaluating our deception detection models. This dataset was gathered from various courtroom trials, ensuring a realistic and challenging environment for deception detection. The dataset comprises three primary components: videos, audio files, and text transcripts.

\subsection{Dataset Composition}
\begin{itemize}
    \item \textbf{Videos:} The dataset contains 121 videos, with 61 labeled as deceptive and 60 labeled as truthful. These videos were recorded during actual courtroom proceedings, capturing a wide range of deceptive and truthful behaviors exhibited by individuals under questioning.
    \item \textbf{Audio:} Each video is accompanied by an audio file in WAV format, resulting in a total of 121 audio files. These audio recordings capture the verbal exchanges and nuances in speech that are critical for detecting deception.
    \item \textbf{Text:} The dataset includes 121 text files, which are transcripts of the spoken content in the videos. These transcripts provide a detailed account of the verbal interactions and are used for linguistic analysis.
    \item \textbf{Frames:} The videos collectively consist of 82,575 frames, which are extracted for visual feature analysis. These frames capture the facial expressions and movements of the individuals, which are essential for identifying non-verbal cues of deception.
\end{itemize}

\section{Experimental Design and Results}

In our study on deception detection, we trained various features, which were already extracted, using different models. We focused on three primary modalities: visual, audio, and text. Each modality was initially trained separately using different models to understand their individual contributions to deception detection. Subsequently, we combined these modalities in pairs and finally, all three together, to evaluate the effectiveness of multi-modal fusion in improving detection accuracy. The dataset was split into 80\% for training and 20\% for testing. All the experiments were on 30 epochs with a 0.0001 learning rate. The following sections detail the methodologies and results for each stage of the experiment.

\begin{figure}[!h]
\vspace{10mm}
\centering
\includegraphics[width=\textwidth]{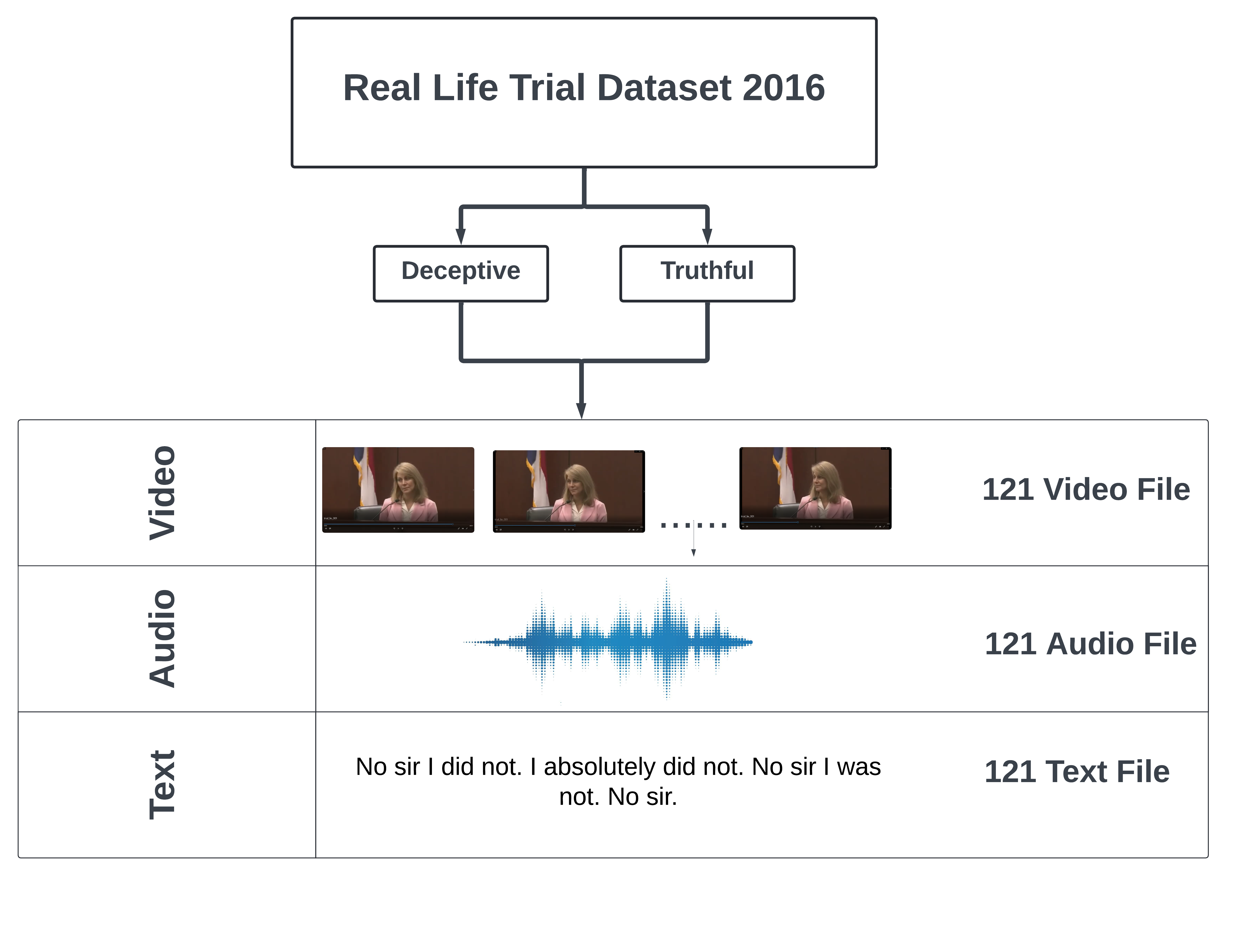}
\caption{Dataset Figure Showing the Dataset.}
\label{fig}
\vspace{10mm}
\end{figure}

\begin{figure}[!h]
\vspace{10mm}
\centering
\includegraphics[width=\textwidth]{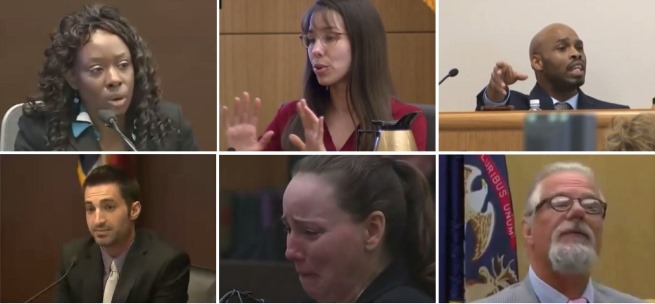}
\caption{Samples from Real life trial Dataset}
\label{fig}
\vspace{10mm}
\end{figure}

\subsection{Single Modality Training}

We trained each modality—visual (V), text (T), and audio (A)—independently to evaluate their individual performance in deception detection. The models used for this stage included LSTM, BiLSTM, VGG16, VGG19, RESNET18, and RESNET50. The results of this training are summarized in Table \ref{tab: table1}. As seen in the table, the LSTM model achieved the highest accuracy for the visual modality (95\%). 

\begin{table}[!h]
\centering
\caption{Results of Voice, Text and Audio alone}
\label{tab: table1}
\begin{tblr}{
  width = \linewidth,
  colspec = {Q[340]Q[150]Q[150]Q[233]},
  hlines,
  vlines,
}
Model    & V    & T    & A       \\
LSTM     & 95\% & 93\% & 88\%    \\
BILSTM   & 90\% & 90\% & 77\%    \\
VGG16    & 85\% & 88\% & 65.70\% \\
VGG19    & 82\% & 86\% & 66\%    \\
RESNET18 & 88\% & 90\% & 75\%    \\
RESNET50 & 90\% & 92\% & 80.00\% 
\end{tblr}
\vspace{10mm}
\end{table}

\subsection{Dual Modality Training}

To investigate the potential benefits of combining modalities, we trained pairs of modalities together: audio and text (A+T), voice and text (V+T), and voice and audio (V+A). The performance of these combinations is presented in Table \ref{tab: table2}. Among the pairs, the voice and text combination (V+T) with the LSTM model showed the highest accuracy (97\%), indicating a significant improvement compared to single modality training.

\begin{table}[!h]
\centering
\caption{Results of different combinations of Voice, Text and Audio}
\label{tab: table2}
\begin{tblr}{
  width = \linewidth,
  colspec = {Q[292]Q[200]Q[200]Q[200]},
  hlines,
  vlines,
}
Model    & A+T     & V+T     & V+A     \\
LSTM     & 90\%    & 97\%    & 87\%    \\
BILSTM   & 85\%    & 90.01\% & 84\%    \\
VGG16    & 79\%    & 92\%    & 77\%    \\
VGG19    & 82\%    & 90\%    & 80\%    \\
RESNET18 & 83.60\% & 88\%    & 83\%    \\
RESNET50 & 83.15\% & 89\%    & 81.30\% 
\end{tblr}
\vspace{10mm}
\end{table}

\subsection{Triple Modality Training}

Finally, we combined all three modalities—voice, audio, and text (V+A+T)—to evaluate the full potential of multi-modal fusion in deception detection. The results, as shown in Table \ref{tab: table3}, reveal that the LSTM model achieved the highest accuracy (99\%) when using all three modalities together. This indicates a substantial enhancement in detection accuracy compared to single and dual modality training.

\begin{table}[!h]
\centering
\caption{Results of a combination of Voice, Audio, and Text all together}
\label{tab: table3}
\begin{tblr}{
  width = \linewidth,
  colspec = {Q[500]Q[319]},
  hlines,
  vlines,
}
Model    & V+A+T \\
LSTM     & 99\%  \\
BILSTM   & 91\%  \\
VGG16    & 92\%  \\
VGG19    & 89\%  \\
RESNET18 & 90\%  \\
RESNET50 & 91\%  
\end{tblr}
\vspace{10mm}
\end{table}

These results underscore the effectiveness of combining multiple modalities in improving deception detection accuracy. The LSTM model consistently outperformed others, especially when using combined modalities. This comprehensive evaluation highlights the potential of multi-modal approaches in the field of deception detection.

\section{Conclusion}

Our research demonstrates the significant potential of multi-modal fusion in enhancing the accuracy of deception detection systems. By combining visual, audio, and text features, we were able to achieve an impressive 99\% accuracy, highlighting the effectiveness of integrating diverse data sources. The study involved a comprehensive analysis of different feature extraction techniques and various machine learning models, including LSTM, BiLSTM, and pre-trained CNNs. The results consistently showed that multi-modal approaches outperform single modality training, underscoring the importance of leveraging multiple data types in deception detection.Furthermore, our findings emphasize the critical role of feature engineering in building robust and interpretable models. By carefully selecting and combining features from different modalities, we were able to capture a wide range of deceptive cues, leading to substantial improvements in detection performance. This approach not only enhances accuracy but also provides a clearer understanding of the underlying mechanisms of deception.
In conclusion, this research provides a solid foundation for future work in the field of automated deception detection. The insights gained from our experiments can be used to develop more sophisticated fusion techniques and incorporate additional modalities, further improving the reliability and transparency of deception detection systems. As the volume of video content continues to grow, the development of advanced multi-modal detection methodologies will be crucial in various applications, from security and law enforcement to social media and beyond.
\vspace{10mm}

\bibliographystyle{plain}
\bibliography{main}

\end{document}